\documentclass[runningheads]{llncs}

\usepackage{graphicx}
\usepackage{subfiles}
\usepackage{comment}
\usepackage{amsmath,amssymb} 
\usepackage{amsfonts}
\usepackage[misc,geometry]{ifsym}
\usepackage{color, soul}
\usepackage{hyperref}
\usepackage{color, colortbl}
\usepackage{multirow}
\usepackage{makecell}
\usepackage{array}

\definecolor{LightCyan}{rgb}{0.88,1,1}

\newif\ifdraft
\drafttrue

\definecolor{orange}{rgb}{1,0.5,0}
\definecolor{pink}{rgb}{0.98, 0.38, 0.5}
\definecolor{darkgreen}{rgb}{0.055, 0.490, 0.016} 

\ifdraft
 \usepackage[normalem]{ulem}
 \newcommand{\RS}[1]{{\color{red}{\bf RS: #1}}}
 
 \newcommand{\PMN}[1]{{\color{orange}{\bf PMN: #1}}}
 
 \newcommand{\STM}[1]{{\color{blue}{\bf STM: #1}}}
 
\else
 \newcommand{\sout}[1]{}
 \newcommand{\RS}[1]{{\color{red}{}}}
 
 \newcommand{\PMN}[1]{{\color{red}{}}}
 
 \newcommand{\STM}[1]{{\color{red}{}}}
 
\fi

\newcommand{\x}{\mathbf{x}}
\newcommand{\p}{\mathbf{p}}
\newcommand{\q}{\mathbf{q}}

\newcommand{\eg}{\emph{e.g.}}
\newcommand{\ie}{\emph{i.e.}}

\begin{document}

\title{Consistency-preserving Visual Question Answering in Medical Imaging}


\titlerunning{Consistency-preserving VQA in Medical Imaging}


\author{Sergio Tascon-Morales \Letter, Pablo Márquez-Neila, Raphael Sznitman}

\authorrunning{Tascon-Morales et al.}

\institute{University of Bern, Bern, Switzerland\\ \email{\{sergio.tasconmorales, pablo.marquez, raphael.sznitman\}@unibe.ch}}

\maketitle          

\begin{abstract}
Visual Question Answering (VQA) models take an image and a natural-language question as input and infer the answer to the question. Recently, VQA systems in medical imaging have gained popularity thanks to potential advantages such as patient engagement and second opinions for clinicians. While most research efforts have been focused on improving architectures and overcoming data-related limitations, answer consistency has been overlooked even though it plays a critical role in establishing trustworthy models. In this work, we propose a novel loss function and corresponding training procedure that allows the inclusion of relations between questions into the training process. Specifically, we consider the case where implications between perception and reasoning questions are known a-priori. To show the benefits of our approach, we evaluate it on the clinically relevant task of Diabetic Macular Edema (DME) staging from fundus imaging. Our experiments show that our method outperforms state-of-the-art baselines, not only by improving model consistency, but also in terms of overall model accuracy. Our code and data are available at \url{https://github.com/sergiotasconmorales/consistency_vqa}.


\end{abstract}
\section{Introduction}
\label{sec:intro}
Visual Question Answering (VQA) models are neural networks that answer natural language questions about an image by interpreting the question and the image provided~\cite{antol2015vqa,goyal2017making,hudson2019gqa,tan2019lxmert}. Specifying questions using natural language gives VQA models great appeal, as the set of possible questions one can ask is enormous and does not need to be identical to the set of questions used to train the models. Due to these advantages, VQA models for medical applications have also been proposed~\cite{gong2021cross,ImageCLEFVQA_Med2018,liao2020aiml,liu2019effective,vu2020question,zhan2020medical}, whereby allowing clinicians to probe the model with subtle differentiating questions and contributing to build trust in predictions.

\begin{figure}[t]
\begin{center}
\includegraphics[width=0.75\textwidth]{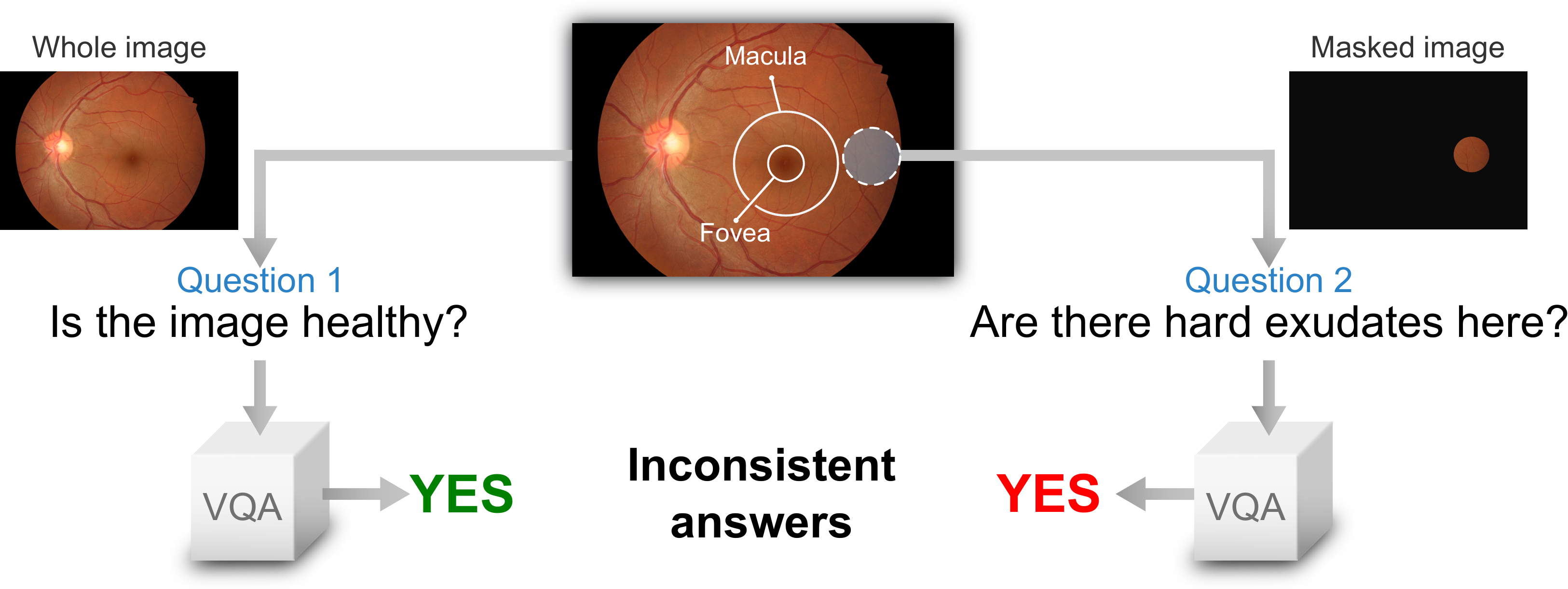}
\caption{VQA inconsistency in Diabetic Macular Edema staging from fundus photograph. While the VQA model correctly answers ``Is the image healthy?" (left), it incorrectly answers yes to``Are there hard exudates here?" for a specified retinal region.}
\label{fig:motivation}
\end{center}
\end{figure}

To date, much of the work in medical VQA has focused on building more effective model architectures~\cite{gong2021cross,liao2020aiml,vu2020question} or overcoming limitations in medical VQA datasets~\cite{Nguyen19,liao2020aiml,sarrouti2020nlm,zhan2020medical}. Yet a critical component of VQA is the notion of {\it consistency} in the answers produced by a model. Here, consistency refers to a model's capacity to produce answers that are not self-contradictory. For instance, the task of staging diabetic macular edema (DME) from color fundus photograph illustrated in Fig.~\ref{fig:motivation} involves identifying {\it perception} elements in the image (\eg,~``are there hard exudates visible near the macula?'') to infer a disease stage, which can be expressed as a {\it reasoning} question (\eg,~``what is the stage of disease?''). Ultimately, for any VQA model to be trustworthy, it should be able to answer these without contradicting itself (\ie, answer that the image is healthy, but also identify hard exudates in the periphery of the eye).

\nocite{wang2021image}

Consistency in VQA has been been studied in the broader computer vision context~\cite{goel2021iq,gokhale2020vqa,ray2019sunny,ribeiro2019red,shah2019cycle}, where the relation between perception and reasoning questions is unconstrained. That is, the answers to perception questions do not necessarily imply any information with respect to the reasoning question and vice-versa. In these broad cases, some methods have modeled question implications \cite{ray2019sunny,ribeiro2019red} or rephrased questions \cite{shah2019cycle} by generating tailored question-answer pairs (\eg,~consistent data-augmentation). Alternatively,~\cite{gokhale2020vqa,teney2019incorporating,yuan2021perception} used relations between questions to impose constraints in the VQA's embedding space. To avoid needing to know the relation between questions,~\cite{selvaraju2020squinting} proposed to enforce consistency by making attention maps of reasoning and perception questions similar to one another. 
However, even though these approaches tackle unconstrained question relations, the ensuring of VQA models' consistency remains limited and often reduces the overall performance~\cite{selvaraju2020squinting}.  

Instead, we propose a novel approach to enforce VQA consistency that is focused on cases where answers to the perception questions have explicit implications on reasoning question answers and vice-versa (\eg, cancerous cells and severity of cancer found in H\&E staining, or presence of hard exudates and DME staging). By focusing on this subset of question relations, our aim is to improve both the accuracy of our model and its consistency, without needing external data as in~\cite{ribeiro2019red,Nguyen19,goel2021iq}. To do this, we allow questions to probe arbitrary image regions by masking irrelevant parts of the image and passing the masked image to the VQA model (see Fig.~\ref{fig:motivation}). To then enforce consistency, we propose a new loss function that penalizes incorrect perceptual predictions when reasoning ones are correct for a given image. To validate the impact of our approach, we test it in the context of DME staging and show that it outperforms state-of-the-art methods for consistency, without compromising overall performance accuracy.

\section{Method}
We present here our approach which consists of using a simple VQA model with a training protocol that encourages consistency among pairs of perception and reasoning questions. Fig.~\ref{fig:method} illustrates this VQA model and our training approach.

\subsubsection{VQA model.} 
\begin{figure}[!t]
\begin{center}
\includegraphics[width=0.70\textwidth]{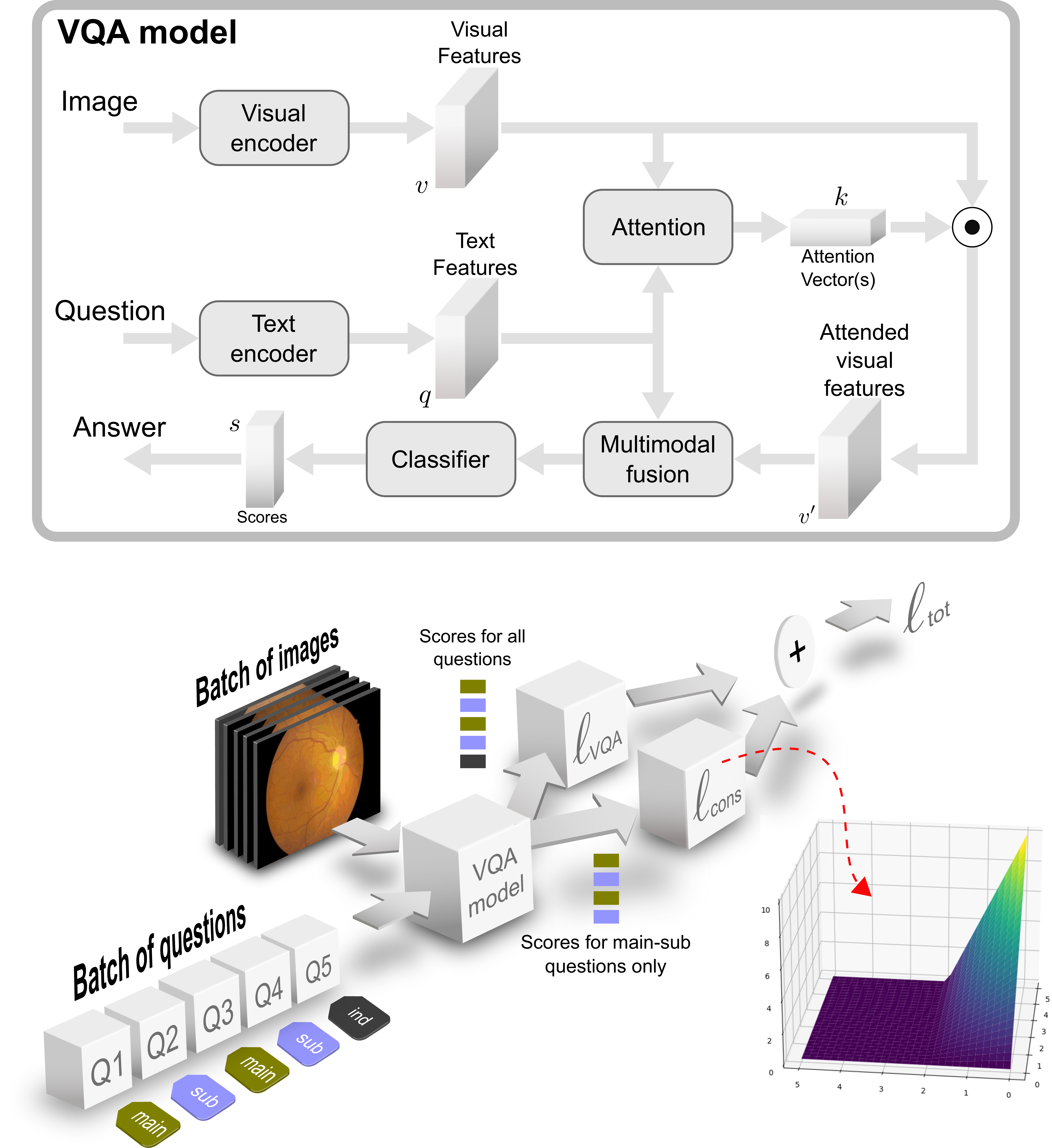}
\caption{\textit{Top:} VQA model architecture. \textit{Bottom:} Visualization of the training process with the proposed loss. The total loss, $\ell_{\textrm{tot}}$, is based on two terms: the conventional VQA loss,  $\ell_{\textrm{VQA}}$ and our proposed consistency loss term, $\ell_{\textrm{cons}}$. The latter is computed only for pairs of main (reasoning) and sub (perception) questions. Training mini-batches consist of main and sub questions at the same time, whereby sub-questions can consider specific regions of the image. Unrelated questions (denoted with ``ind") can also be included in training batches, but do not contribute to $\ell_{\textrm{cons}}$.}
\label{fig:method}
\end{center}
\end{figure}

Following~\cite{cadene2019rubi}, our VQA model,~$f:\mathcal{I}\times\mathcal{Q}\to \mathcal{P}(\mathcal{A})$, takes a tuple containing an image,~$\x$, and a question,~$\q$, to produce a distribution,~$\p=f(\x,\q)$, over a finite set of possible answers~$\mathcal{A}$ (see Fig.~\ref{fig:method}(Top)). After encoding the inputs, the VQA model combines visual ($v$) and textual ($q$) features through an attention module ($k$)~\cite{xu2015show} that selects the visual features relevant to the question ($v'$). The final classifier receives a combination of the relevant features and the text features through a fusion module to predict the final distribution.

In some cases, questions may consider asking about content related to specific regions of the image (\eg, ``are there hard exudates in this region?''). To process these cases, the input image is masked so that the visible area corresponds to the region mentioned in the question while the rest of the image is set to zero.

Training this model requires a dataset~$\mathcal{T}=\{t^{(i)}=(\x^{(i)}, \q^{(i)}, a^{(i)})\}_{i=1}^N\subseteq\mathcal{I}\times\mathcal{Q}\times\mathcal{A}$ of images and questions annotated with their answers. The VQA~loss is simply the cross-entropy between the predicted distribution and the real answer,
\begin{equation}
    \label{eq:vqa_loss}
    \ell_{\textrm{VQA}}(\p, a) = H(\p, a) = -\log \p_a.
\end{equation}
While this loss alone is sufficient to reach a reasonable performance, it ignores the potentially useful interactions that may exist among training questions.

\subsubsection{Consistency loss.} We aim to improve the quality of our VQA model by exploiting the relationships between reasoning and perception questions at training time. To this end, we augment the training dataset with an additional binary relation~$\prec$ over the set of questions~$\mathcal{Q}$. Two questions are related, $\q^{(i)}\prec \q^{(j)}$, if $\q^{(i)}$ is a perception question associated to the reasoning question~$\q^{(j)}$. From hence on, we refer to perception questions as \emph{sub-questions} and reasoning questions as \emph{main questions}.

Following the terminology in~\cite{selvaraju2020squinting}, an inconsistency occurs when the VQA model  infers the main question correctly but the sub-question incorrectly. Using the entropy as a measurement of incorrectness, we propose to impose the consistency at training time by penalizing the cases with high $H^{(i)}=H(\p^{(i)}, a^{(i)})$ and low $H^{(j)}=H(\p^{(j)}, a^{(j)})$ when $\q^{(i)}\prec \q^{(j)}$. To do this, we use an adapted hinge loss that disables the penalty when $H^{(j)}$ is larger than a threshold~$\gamma>0$, but otherwise penalizes large values of~$H^{(i)}$,
\begin{equation}
    \label{eq:cons_loss}
    \ell_{\textrm{cons}}(H^{(i)}, H^{(j)}) = H^{(i)}\max\{0, \gamma - H^{(j)}\}.
\end{equation}
\noindent

The final cost function then minimizes the expected value of the VQA loss~\eqref{eq:vqa_loss} for the elements of the training dataset and the consistency loss~\eqref{eq:cons_loss} for the pairs of training samples with $\prec$-related questions,
\begin{equation}
    \label{eq:cost}
    \mathbb{E}_{t\sim\mathcal{T}}[\ell_\textrm{VQA}(\p, a)] +
    \lambda\mathbb{E}_{(t^{(i)}, t^{(j)})\sim\mathcal{T}^2}[\ell_\textrm{cons}(H^{(i)}, H^{(j)}) \mid \x^{(i)}=\x^{(j)}, \q^{(i)}\prec\q^{(j)}],
\end{equation}
where $\lambda > 0$ controls the relative strength of both losses and $\mathcal{T}^2$ is the Cartesian product of $\mathcal{T}$ with itself, that is, all pairs of training samples.

To train, this cost is iteratively minimized approximating the expectations with mini-batches. The two expectations of Eq.~\eqref{eq:cost} suggest that two mini-batches are necessary at each iteration: one mini-batch sampled from~$\mathcal{T}$ and a second mini-batch of $\prec$-related pairs sampled from $\mathcal{T}^2$. However, in practice a single mini-batch is sufficient as long as we ensure that it contains pairs of $\prec$-related questions. While this biased sampling could in turn bias the estimation of the first expectation, we did not observe a noticeable impact in our experiments. Fig.~\ref{fig:method}(Bottom) illustrates this training procedure.

\section{Experiments and results}

\subsubsection{DME staging.}
\begin{figure}[b!]
\begin{center}
\includegraphics[width=0.9\textwidth]{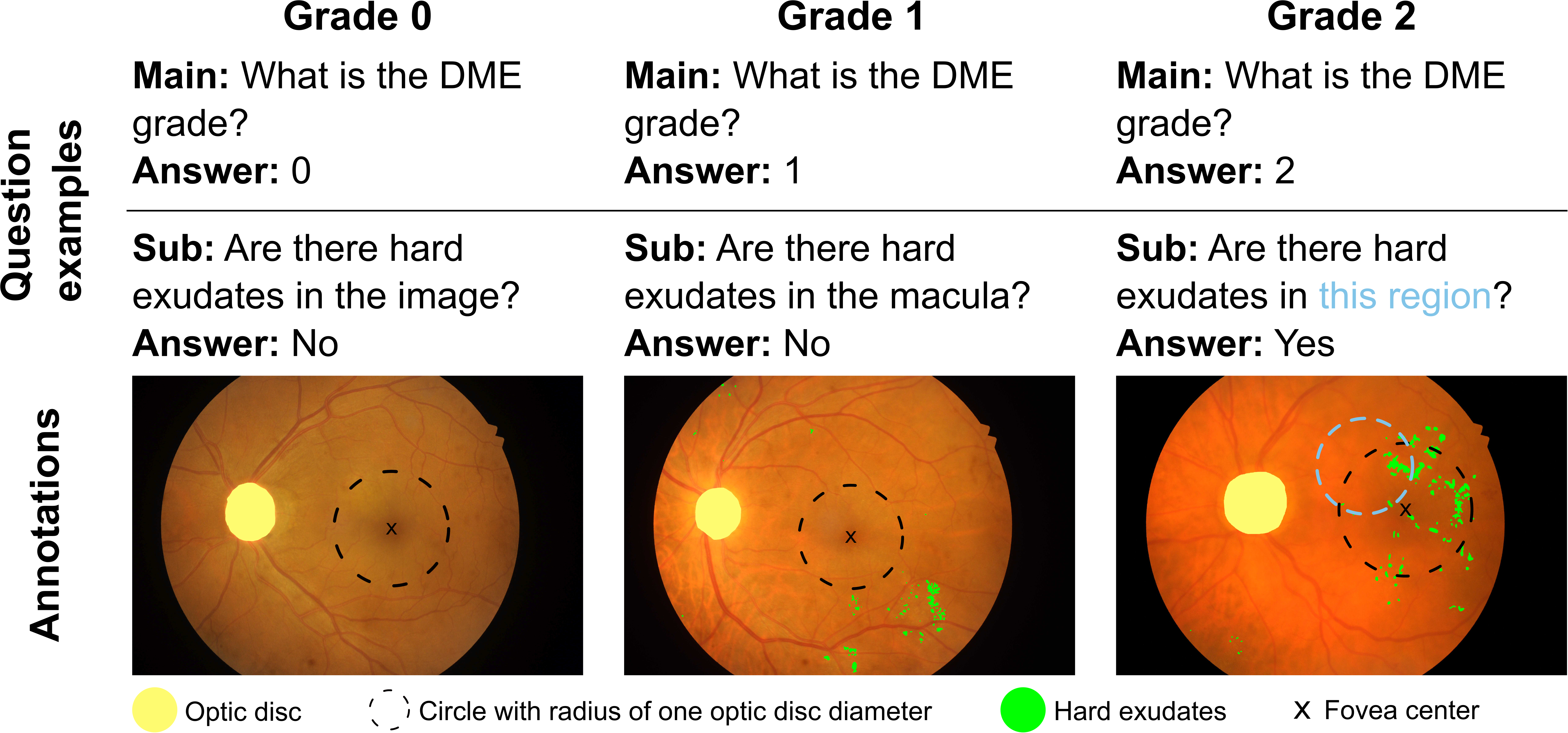}
\caption{DME risk grading Grade 0 is assigned if there are no hard exudates present in the whole image. Grade 1 is assigned if there are hard exudates, but only located outside a circle centered at the fovea with radius of one optic disc diameter. Grade 2 is assigned if there are hard exudates located within the circle. Examples of main and sub-questions are provided for each grade.}
\label{fig:dme}
\end{center}
\end{figure}

Diabetic Macular Edema (DME) staging from color fundus images involves grading images on a scale from 0 to 2, with 0 being healthy and 2 being severe (see Fig.~\ref{fig:dme}). Differentiation between the grades relies on the presence of hard exudates present in different locations of the retina. Specifically, a grade of 0 implies that no hard exudates are present at all, a grade of 1 implies that hard exudates are located in the retina periphery (\ie,~outside a circular region centered at the fovea center with radius of one optic disc diameter), and a grade of 2 when hard exudates are in the macular region~\cite{ren2018diabetic}.

\subsubsection{Dataset and questions.}
To validate our method, we make use of two publicly available datasets: the Indian Diabetic Retinopathy Image Dataset (IDRiD)~\cite{idrid} and the e-Ophta dataset~\cite{decenciere2013teleophta}. From the IDRiD dataset, we use images from the segmentation and grading tasks, which consist of 81 and 516 images, respectively. Images from the segmentation task include segmentation masks for hard exudates and images from the grading task only have the DME grade. On the other hand, the e-Ophta dataset comprises 47 images with segmentation of hard exudates and 35 images without lesions. Combining both datasets yields a dataset of 128 images with segmentation masks for hard exudates and 128 images without any lesions, plus 423 images for which only the DME risk grade is available. 

In this context, we consider main questions to be those asking ``What is the DME risk grade?" when considering the entire image. Sub-questions were then defined as questions asking about the presence of the hard exudates. For instance, as shown in Fig.~\ref{fig:dme}{(Right)}, ``Are there hard exudates in this region?" where the region designated contains the macula. In practice, we set three types of sub-questions: ``are there hard exudates in this image?", ``are there hard exudates in the macula?" and ``are there hard exudates in this region?". We refer to these three questions as \textit{whole}, \textit{macula} and \textit{region} questions, respectively. For the region sub-questions, we consider circular regions that can be centered anywhere, or centered on the fovea, depending on availability of fovea center location annotations. As mentioned in Sec.~\ref{sec:intro}, to answer questions about regions, images are masked so that only the region is visible.

The total number of question-answer pairs in our dataset consist of 9779 for training {(4.4\% main, 21.4\% sub, 74.2\% ind)}, 2380 for validation {(4.5\% main, 19.2\% sub, 76.3\% ind)} and 1311 for testing {(10\% main, 46.1\% sub, 43.9\% ind)}, with images in the train, validation and test sets being mutually exclusive.

\subsubsection{Baselines, implementation details and evaluation metrics:}
We compare our approach to a baseline model that does not use the proposed $\ell_{\textrm{cons}}$ loss, equivalent to setting $\lambda=0$. In addition, we compare our method against the attention-matching method, SQuINT~\cite{selvaraju2020squinting}, as it is a state-of-the-art alternative to our approach that can be used with the same VQA model architecture.

Our VQA model uses an ImageNet-pretrained ResNet101~\cite{he2016deep} with input image of $448\times 448$~pixels and an embedding of 2048~dimensions for the image encoding. For text encoding, a single-layer LSTM~\cite{hochreiter1997long} network processes the input question with word encoding of length 300 and produces a single question embedding of 1024~dimensions. The multi-glimpse attention mechanism~\cite{xu2015show} uses 2~glimpses and dropout rate~$0.25$, and the multimodal fusion stage uses standard concatenation. The final classifier is a multi-layer perceptron with hidden layer of 1024 dimensions and dropout rate of 0.25. Hyperparameters~$\lambda$ and~$\gamma$ were empirically adjusted to 0.5 and~1.0, respectively. 

All experiments were implemented using PyTorch~1.10.1 and run on a Linux machine with an NVIDIA RTX 3090 graphic card using 16~GB of memory and 4~CPU cores. All methods use the weighted cross-entropy as the base VQA loss function. Batch size was set to~64, and we used Adam for optimization with a learning rate of $10^{-4}$. Maximum epoch number was 100 and we use early stopping policy to prevent overfitting, with a patience of 20~epochs. 

We report accuracy and consistency~\cite{selvaraju2020squinting} performances, using two different definitions of consistency for comparison. Consistency, C1, is the percentage of sub-questions that are answered correctly when the main question was answered correctly. Consistency, C2, is the percentage of main questions that are answered correctly when all corresponding sub-questions were answered correctly.

\subsubsection{Results:}
\begin{figure}[!t]
\begin{center}
\includegraphics[width=0.99\textwidth]{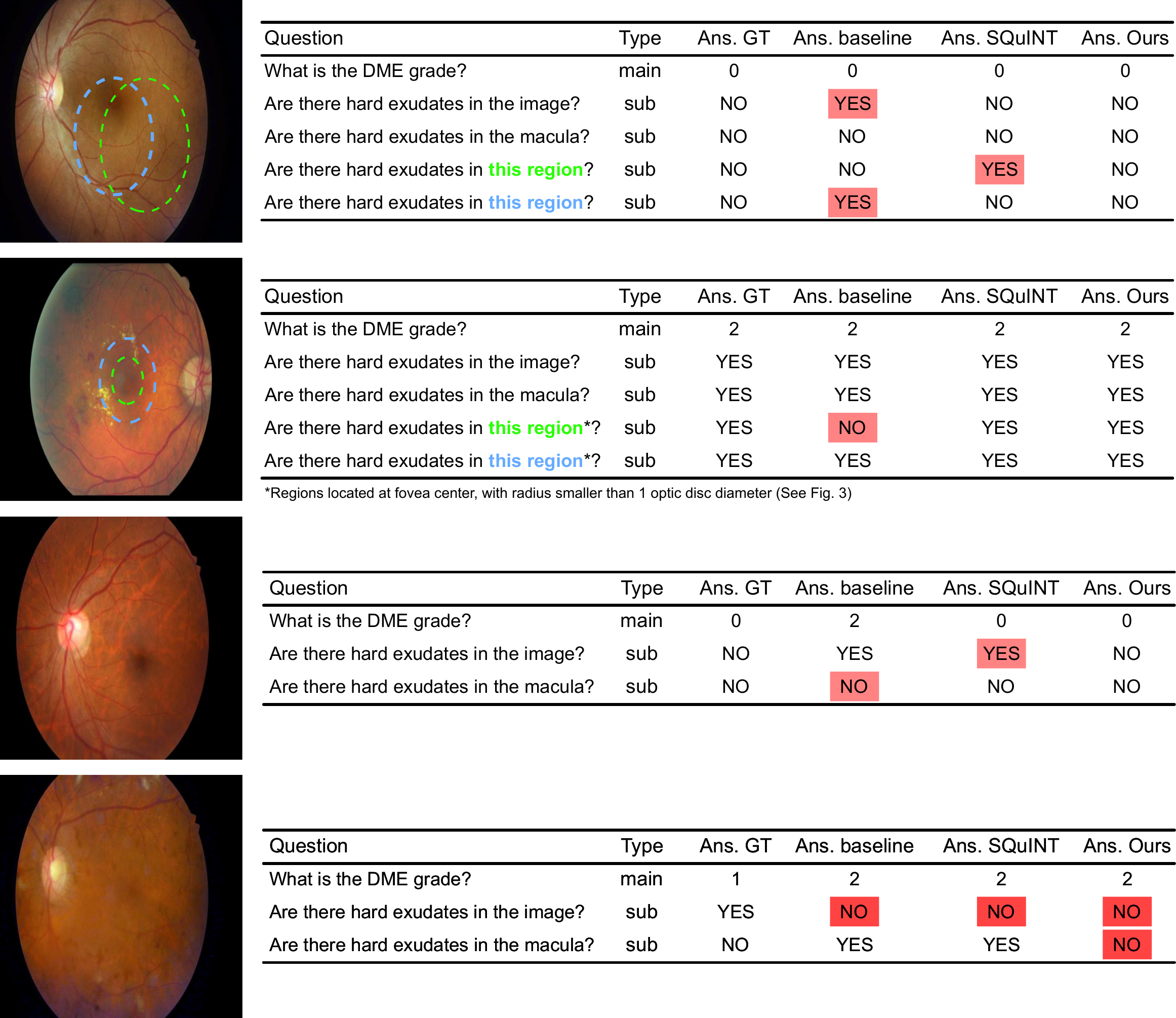}
\caption{Qualitative examples from the test set. Inconsistent sub-answers are highlighted in red. Additional examples are shown in the supplementary material.  
}
\label{fig:examples}
\end{center}
\end{figure}

Table \ref{tab:results} shows the results.
We compare these results to the case in which the value of $\lambda$ is~0, which corresponds to the baseline in which no additional loss term is used. For each case, we present the overall accuracy and the accuracy for each type of question, as well as the consistency values. Fig.~\ref{fig:examples} illustrates the performance of each method with representative qualitative examples. 
\begin{table}[!t]
\begin{center}
\resizebox{\textwidth}{!}{
\begin{tabular}{llllllll}
\hline
\multicolumn{1}{|l|}{\multirow{2}{*}{Case}}    &  \multicolumn{5}{c|}{Accuracy} & \multicolumn{2}{c|}{Consistency} \\ \cline{2-8} \multicolumn{1}{|c|}{}  
                   & \multicolumn{1}{l|}{overall}      & \multicolumn{1}{l|}{grade}        & \multicolumn{1}{l|}{whole}        & \multicolumn{1}{l|}{macula}       & \multicolumn{1}{l|}{region}       & \multicolumn{1}{c|}{C1}            & \multicolumn{1}{c|}{C2}          

\\ \hline
\multicolumn{1}{|l|}{Baseline (no att.)}                                 & \multicolumn{1}{l|}{77.54 } & \multicolumn{1}{l|}{73.59} & \multicolumn{1}{l|}{81.37 } & \multicolumn{1}{l|}{83.37}& \multicolumn{1}{l|}{76.66 } & \multicolumn{1}{l|}{81.70 }  & \multicolumn{1}{l|}{91.86 } 

\\ \hline
\multicolumn{1}{|l|}{Baseline (att.)}                            & \multicolumn{1}{l|}{81.46 } & \multicolumn{1}{l|}{80.23} & \multicolumn{1}{l|}{83.13 } & \multicolumn{1}{l|}{\textbf{87.18} }& \multicolumn{1}{l|}{80.58 } & \multicolumn{1}{l|}{89.21 }  & \multicolumn{1}{l|}{96.92 } 

\\ \hline
\multicolumn{1}{|l|}{Baseline (att.) + SQuINT~\cite{selvaraju2020squinting}  }                                  & \multicolumn{1}{l|}{80.58} & \multicolumn{1}{l|}{77.48} & \multicolumn{1}{l|}{82.82} & \multicolumn{1}{l|}{85.34}& \multicolumn{1}{l|}{80.02} & \multicolumn{1}{l|}{88.17}  & \multicolumn{1}{l|}{94.62} 

\\ \hline
\multicolumn{1}{|l|}{Baseline (att.) + Ours ($\lambda=0.5,\gamma=1$)}             & \multicolumn{1}{l|}{\textbf{83.49} } & \multicolumn{1}{l|}{\textbf{80.69} } & \multicolumn{1}{l|}{\textbf{84.96} } & \multicolumn{1}{l|}{\textbf{87.18} } & \multicolumn{1}{l|}{\textbf{83.16}} & \multicolumn{1}{l|}{\textbf{94.20} } & \multicolumn{1}{l|}{\textbf{98.12} } 

\\ \hline                  
\end{tabular}
}
\end{center}
\caption{Average test accuracy and consistency values for the different models. Results shown are averaged over 10 models trained with different seeds. Accuracy values are presented for all questions (overall), for main questions (grade) and for sub-questions (whole, macula and region). Both measures of consistency are shown as well.}
\label{tab:results}
\end{table}

\begin{table}[!t]
\begin{center}
\begin{tabular}{lllllllll}
\hline
 \multicolumn{1}{|c|}{\multirow{2}{*}{$\lambda$}} & \multicolumn{1}{c|}{\multirow{2}{*}{$\gamma$}} & \multicolumn{5}{c|}{Accuracy}   & \multicolumn{2}{c|}{Consistency}  \\ \cline{3-9} 
\multicolumn{1}{|c|}{}                         & \multicolumn{1}{c|}{}      &  \multicolumn{1}{c|}{overall}                       & \multicolumn{1}{c|}{grade}      & \multicolumn{1}{c|}{whole}        & \multicolumn{1}{c|}{macula}        & \multicolumn{1}{c|}{region}     & \multicolumn{1}{c|}{C1}            & \multicolumn{1}{c|}{C2}          
\\ \hline

\multicolumn{1}{|l|}{0}                     & \multicolumn{1}{l|}{-}                    & \multicolumn{1}{l|}{81.46} & \multicolumn{1}{l|}{80.23} & \multicolumn{1}{l|}{83.13} & \multicolumn{1}{l|}{87.18} & \multicolumn{1}{l|}{80.58} & \multicolumn{1}{l|}{89.21} & \multicolumn{1}{l|}{96.92} 

\\ \hline \hline

\multicolumn{1}{|l|}{0.2}                     & \multicolumn{1}{l|}{0.5}                    & \multicolumn{1}{l|}{82.01} & \multicolumn{1}{l|}{80.38} & \multicolumn{1}{l|}{83.59} & \multicolumn{1}{l|}{86.56} & \multicolumn{1}{l|}{81.36} & \multicolumn{1}{l|}{90.93} & \multicolumn{1}{l|}{97.38} 

\\ \hline
\multicolumn{1}{|l|}{0.2}                     & \multicolumn{1}{l|}{1}                    & \multicolumn{1}{l|}{82.65} & \multicolumn{1}{l|}{ 79.77} & \multicolumn{1}{l|}{ 83.97} & \multicolumn{1}{l|}{ 86.64} & \multicolumn{1}{l|}{82.30} & \multicolumn{1}{l|}{ 93.22} & \multicolumn{1}{l|}{97.51 } 

\\ \hline

\multicolumn{1}{|l|}{0.2}                     & \multicolumn{1}{l|}{1.5}                    & \multicolumn{1}{l|}{83.05} & \multicolumn{1}{l|}{81.22} & \multicolumn{1}{l|}{ 84.27} & \multicolumn{1}{l|}{ 87.33} & \multicolumn{1}{l|}{82.53} & \multicolumn{1}{l|}{ 93.23} & \multicolumn{1}{l|}{ 97.56}

\\ \hline \hline
\multicolumn{1}{|l|}{0.3}                     & \multicolumn{1}{l|}{0.5}                   & \multicolumn{1}{l|}{82.34} & \multicolumn{1}{l|}{ 79.92} & \multicolumn{1}{l|}{ 83.59} & \multicolumn{1}{l|}{ 87.71} & \multicolumn{1}{l|}{81.74} & \multicolumn{1}{l|}{ 92.32} & \multicolumn{1}{l|}{ 97.31} 

\\ \hline

\multicolumn{1}{|l|}{0.3}                     & \multicolumn{1}{l|}{1}                 & \multicolumn{1}{l|}{83.27} & \multicolumn{1}{l|}{ 80.53} & \multicolumn{1}{l|}{ 84.58} & \multicolumn{1}{l|}{ 87.25} & \multicolumn{1}{l|}{82.91} & \multicolumn{1}{l|}{ 94.01} & \multicolumn{1}{l|}{ 98.10} 

\\ \hline

\multicolumn{1}{|l|}{0.3}                     & \multicolumn{1}{l|}{1.5}                   & \multicolumn{1}{l|}{83.28} & \multicolumn{1}{l|}{80.84 } & \multicolumn{1}{l|}{ 84.43} & \multicolumn{1}{l|}{ 87.48} & \multicolumn{1}{l|}{82.86} & \multicolumn{1}{l|}{ 93.28} & \multicolumn{1}{l|}{ 98.29}

\\ \hline \hline

\multicolumn{1}{|l|}{0.4}                     & \multicolumn{1}{l|}{0.5}                   & \multicolumn{1}{l|}{82.87} & \multicolumn{1}{l|}{80.69} & \multicolumn{1}{l|}{ 84.89} & \multicolumn{1}{l|}{ 87.02} & \multicolumn{1}{l|}{82.30} & \multicolumn{1}{l|}{ 92.66} & \multicolumn{1}{l|}{ 96.66} 

\\ \hline
\multicolumn{1}{|l|}{0.4}                     & \multicolumn{1}{l|}{1}                 & \multicolumn{1}{l|}{82.97} & \multicolumn{1}{l|}{ 80.15} & \multicolumn{1}{l|}{ 83.97} & \multicolumn{1}{l|}{ 86.72} & \multicolumn{1}{l|}{82.69} & \multicolumn{1}{l|}{ 93.91} & \multicolumn{1}{l|}{98.23} 

\\ \hline
\multicolumn{1}{|l|}{0.4}                     & \multicolumn{1}{l|}{1.5}                  & \multicolumn{1}{l|}{83.33} & \multicolumn{1}{l|}{ 80.08} & \multicolumn{1}{l|}{ 84.20} & \multicolumn{1}{l|}{ 86.87} & \multicolumn{1}{l|}{83.17} & \multicolumn{1}{l|}{ 93.96} & \multicolumn{1}{l|}{ 97.77} 

\\ \hline \hline
\multicolumn{1}{|l|}{0.5}                     & \multicolumn{1}{l|}{0.5}                & \multicolumn{1}{l|}{82.54} & \multicolumn{1}{l|}{81.07 } & \multicolumn{1}{l|}{ 83.66} & \multicolumn{1}{l|}{ 88.02} & \multicolumn{1}{l|}{81.81} & \multicolumn{1}{l|}{ 91.87} & \multicolumn{1}{l|}{ 97.73} 

\\ \hline
\multicolumn{1}{|l|}{0.5}                     & \multicolumn{1}{l|}{1}                    & \multicolumn{1}{l|}{83.49} & \multicolumn{1}{l|}{80.69 } & \multicolumn{1}{l|}{84.96 } & \multicolumn{1}{l|}{87.18} & \multicolumn{1}{l|}{83.16} & \multicolumn{1}{l|}{94.20} & \multicolumn{1}{l|}{98.12 } 

\\ \hline
\multicolumn{1}{|l|}{0.5}                     & \multicolumn{1}{l|}{1.5}             & \multicolumn{1}{l|}{83.25} & \multicolumn{1}{l|}{ 79.92} & \multicolumn{1}{l|}{ 84.58} & \multicolumn{1}{l|}{ 86.95} & \multicolumn{1}{l|}{83.01} & \multicolumn{1}{l|}{ 94.20} & \multicolumn{1}{l|}{ 98.12} 

\\ \hline                  
\end{tabular}
\end{center}
\caption{Average test accuracy and consistency values for different values of the parameters $\lambda$ and $\gamma$. The first row ($\lambda\, =\, 0$) corresponds to the baseline.}
\label{tab:results_params}
\end{table}
In general, we observe that our proposed approach yields increases in accuracy and consistency when compared to both the baseline and SQuINT. Importantly, this increase in consistency is not at the expense of overall accuracy. Specifically, this indicates that our loss term causes the model to be correct about sub-questions when it is correct about main questions. The observed increase in accuracy also indicates that our approach is not synthetically increasing consistency by reducing the number of correct answers on main questions~\cite{selvaraju2020squinting}. We note that SQuINT results in a reduction in accuracy and consistency, which can be partially explained by the presence of region questions that are not associated to any main question. These questions, which exceed the number of main questions, may affect the constraint in the learned attention maps.

Table~\ref{tab:results_params} shows the effect of $\lambda$ and $\gamma$ on the performance metrics. As expected, we notice that when $\lambda$ increases, the consistency of our approach increases as well and will occasionally deteriorate overall accuracy. The impact of $\gamma$ however is less evident, as no clear trend is visible. This would imply that the exact parameter value used is moderately critical to performances. 

\section{Conclusions}
\label{sec:conc}
In this work, we presented a novel method for improving consistency in VQA models in cases where answers to sub-questions imply those of main questions and vice-versa. By using a tailored training procedure and loss function that measures the level of inconsistency, we show on the application of DME staging, that our approach provides important improvements in both VQA accuracy and consistency. In addition, we show that our method's hyperparameters are relatively insensitive to model performance. In the future, we plan to investigate how this approach can be extended to the broader case of unconstrained question relations.

\subsubsection{Acknowledgments.} This work was partially funded by the Swiss National Science Foundation through the grant \# 191983.


\bibliographystyle{splncs04}
\bibliography{mybibliography}

\begin{thebibliography}{10}
\providecommand{\url}[1]{\texttt{#1}}
\providecommand{\urlprefix}{URL }
\providecommand{\doi}[1]{https://doi.org/#1}

\bibitem{antol2015vqa}
Antol, S., Agrawal, A., Lu, J., Mitchell, M., Batra, D., Zitnick, C.L., Parikh,
  D.: Vqa: Visual question answering. In: Proceedings of the IEEE international
  conference on computer vision. pp. 2425--2433 (2015)

\bibitem{cadene2019rubi}
Cadene, R., Dancette, C., Cord, M., Parikh, D., et~al.: Rubi: Reducing unimodal
  biases for visual question answering. Advances in neural information
  processing systems  \textbf{32},  841--852 (2019)

\bibitem{decenciere2013teleophta}
Decenciere, E., Cazuguel, G., Zhang, X., Thibault, G., Klein, J.C., Meyer, F.,
  Marcotegui, B., Quellec, G., Lamard, M., Danno, R., et~al.: Teleophta:
  Machine learning and image processing methods for teleophthalmology. Irbm
  \textbf{34}(2),  196--203 (2013)

\bibitem{goel2021iq}
Goel, V., Chandak, M., Anand, A., Guha, P.: {IQ-VQA}: Intelligent visual
  question answering. In: International Conference on Pattern Recognition. pp.
  357--370. Springer (2021)

\bibitem{gokhale2020vqa}
Gokhale, T., Banerjee, P., Baral, C., Yang, Y.: {VQA-LOL}: Visual question
  answering under the lens of logic. In: European conference on computer
  vision. pp. 379--396. Springer (2020)

\bibitem{gong2021cross}
Gong, H., Chen, G., Liu, S., Yu, Y., Li, G.: Cross-modal self-attention with
  multi-task pre-training for medical visual question answering. In:
  Proceedings of the 2021 International Conference on Multimedia Retrieval. pp.
  456--460 (2021)

\bibitem{goyal2017making}
Goyal, Y., Khot, T., Summers-Stay, D., Batra, D., Parikh, D.: Making the v in
  vqa matter: Elevating the role of image understanding in visual question
  answering. In: Proceedings of the IEEE Conference on Computer Vision and
  Pattern Recognition. pp. 6904--6913 (2017)

\bibitem{ImageCLEFVQA_Med2018}
Hasan, S.A., Ling, Y., Farri, O., Liu, J., Lungren, M., M\"uller, H.: Overview
  of the {ImageCLEF} 2018 medical domain visual question answering task. In:
  CLEF2018 Working Notes. {CEUR} Workshop Proceedings, CEUR-WS.org
  $<$http://ceur-ws.org$>$, Avignon, France (September 10-14 2018)

\bibitem{he2016deep}
He, K., Zhang, X., Ren, S., Sun, J.: Deep residual learning for image
  recognition. In: Proceedings of the IEEE conference on computer vision and
  pattern recognition. pp. 770--778 (2016)

\bibitem{hochreiter1997long}
Hochreiter, S., Schmidhuber, J.: Long short-term memory. Neural computation
  \textbf{9}(8),  1735--1780 (1997)

\bibitem{hudson2019gqa}
Hudson, D.A., Manning, C.D.: Gqa: a new dataset for compositional question
  answering over real-world images. arXiv preprint arXiv:1902.09506
  \textbf{3}(8) (2019)

\bibitem{liao2020aiml}
Liao, Z., Wu, Q., Shen, C., Van Den~Hengel, A., Verjans, J.: Aiml at vqa-med
  2020: Knowledge inference via a skeleton-based sentence mapping approach for
  medical domain visual question answering  (2020)

\bibitem{liu2019effective}
Liu, F., Peng, Y., Rosen, M.P.: An effective deep transfer learning and
  information fusion framework for medical visual question answering. In:
  International Conference of the Cross-Language Evaluation Forum for European
  Languages. pp. 238--247. Springer (2019)

\bibitem{Nguyen19}
Nguyen, B.D., Do, T.T., Nguyen, B.X., Do, T., Tjiputra, E., Tran, Q.D.:
  Overcoming data limitation in medical visual question answering. In: Shen,
  D., Liu, T., Peters, T.M., Staib, L.H., Essert, C., Zhou, S., Yap, P.T.,
  Khan, A. (eds.) Medical Image Computing and Computer Assisted Intervention --
  MICCAI 2019. pp. 522--530. Springer International Publishing, Cham (2019)

\bibitem{idrid}
Porwal, P., Pachade, S., Kamble, R., Kokare, M., Deshmukh, G., Sahasrabuddhe,
  V., Meriaudeau, F.: Indian diabetic retinopathy image dataset (idrid) (2018).
  \doi{10.21227/H25W98}, \url{https://dx.doi.org/10.21227/H25W98}

\bibitem{ray2019sunny}
Ray, A., Sikka, K., Divakaran, A., Lee, S., Burachas, G.: Sunny and dark
  outside?! improving answer consistency in vqa through entailed question
  generation. arXiv preprint arXiv:1909.04696  (2019)

\bibitem{ren2018diabetic}
Ren, F., Cao, P., Zhao, D., Wan, C.: Diabetic macular edema grading in retinal
  images using vector quantization and semi-supervised learning. Technology and
  Health Care  \textbf{26}(S1),  389--397 (2018)

\bibitem{ribeiro2019red}
Ribeiro, M.T., Guestrin, C., Singh, S.: Are red roses red? evaluating
  consistency of question-answering models. In: Proceedings of the 57th Annual
  Meeting of the Association for Computational Linguistics. pp. 6174--6184
  (2019)

\bibitem{sarrouti2020nlm}
Sarrouti, M.: Nlm at vqa-med 2020: Visual question answering and generation in
  the medical domain. In: CLEF (Working Notes) (2020)

\bibitem{selvaraju2020squinting}
Selvaraju, R.R., Tendulkar, P., Parikh, D., Horvitz, E., Ribeiro, M.T., Nushi,
  B., Kamar, E.: Squinting at vqa models: Introspecting vqa models with
  sub-questions. In: Proceedings of the IEEE/CVF Conference on Computer Vision
  and Pattern Recognition. pp. 10003--10011 (2020)

\bibitem{shah2019cycle}
Shah, M., Chen, X., Rohrbach, M., Parikh, D.: Cycle-consistency for robust
  visual question answering. In: Proceedings of the IEEE/CVF Conference on
  Computer Vision and Pattern Recognition. pp. 6649--6658 (2019)

\bibitem{tan2019lxmert}
Tan, H., Bansal, M.: Lxmert: Learning cross-modality encoder representations
  from transformers. arXiv preprint arXiv:1908.07490  (2019)

\bibitem{teney2019incorporating}
Teney, D., Abbasnejad, E., Hengel, A.v.d.: On incorporating semantic prior
  knowledge in deep learning through embedding-space constraints. arXiv
  preprint arXiv:1909.13471  (2019)

\bibitem{vu2020question}
Vu, M.H., L{\"o}fstedt, T., Nyholm, T., Sznitman, R.: A question-centric model
  for visual question answering in medical imaging. IEEE transactions on
  medical imaging  \textbf{39}(9),  2856--2868 (2020)

\bibitem{wang2021image}
Wang, P., Liao, R., Moyer, D., Berkowitz, S., Horng, S., Golland, P.: Image
  classification with consistent supporting evidence. In: Machine Learning for
  Health. pp. 168--180. PMLR (2021)

\bibitem{xu2015show}
Xu, K., Ba, J., Kiros, R., Cho, K., Courville, A., Salakhudinov, R., Zemel, R.,
  Bengio, Y.: Show, attend and tell: Neural image caption generation with
  visual attention. In: International conference on machine learning. pp.
  2048--2057. PMLR (2015)

\bibitem{yuan2021perception}
Yuan, Y., Wang, S., Jiang, M., Chen, T.Y.: Perception matters: Detecting
  perception failures of vqa models using metamorphic testing. In: Proceedings
  of the IEEE/CVF Conference on Computer Vision and Pattern Recognition. pp.
  16908--16917 (2021)

\bibitem{zhan2020medical}
Zhan, L.M., Liu, B., Fan, L., Chen, J., Wu, X.M.: Medical visual question
  answering via conditional reasoning. In: Proceedings of the 28th ACM
  International Conference on Multimedia. pp. 2345--2354 (2020)

\end{thebibliography}

\end{document}